\def\year{2022}\relax
\documentclass[letterpaper]{article} 
\usepackage{aaai22}  
\usepackage{times}  
\usepackage{helvet}  
\usepackage{courier}  
\usepackage[hyphens]{url}  
\usepackage{graphicx} 
\usepackage{natbib}  
\usepackage{caption} 
\DeclareCaptionStyle{ruled}{labelfont=normalfont,labelsep=colon,strut=off} 
\usepackage{algorithm}
\usepackage{algorithmic}
\usepackage{verbatim}
\usepackage{newfloat}
\usepackage{listings}
\floatstyle{ruled}
\newfloat{listing}{tb}{lst}{}
\floatname{listing}{Listing}
\title{Anti-Money Laundering Alert Optimization Using Machine Learning with Graphs}
\author {
    Ahmad Naser Eddin \textsuperscript{\rm 1},
    Jacopo Bono \textsuperscript{\rm 1},
    David Aparício \textsuperscript{\rm 2},
    David Polido \textsuperscript{\rm 1}, \\
    João Tiago Ascensão \textsuperscript{\rm 1},
    Pedro Bizarro \textsuperscript{\rm 1},
    Pedro Ribeiro \textsuperscript{\rm 2}
}
\affiliations {
    \textsuperscript{\rm 1} Feedzai, Portugal\\
    \textsuperscript{\rm 2} DCC-FCUP, University of Porto, Portugal\\ 
    ahmad.eddin@feedzai.com, 
    jacopo.bono@feedzai.com,
    daparicio@dcc.fc.up.pt,
    david.polido@feedzai.com,
    joao.ascensao@feedzai.com, pedro.bizarro@feedzai.com,
    pribeiro@dcc.fc.up.pt
}

\usepackage{multirow}
\usepackage{array}
\newcolumntype{P}[1]{>{\centering\arraybackslash}m{#1}}

\begin{document}

\maketitle

\begin{abstract}

Money laundering is a global problem that concerns legitimizing proceeds from serious felonies (€1.7-4 trillion annually) such as drug dealing, human trafficking, or corruption. The anti-money laundering systems deployed by financial institutions typically comprise rules aligned with regulatory frameworks. Human investigators review the alerts and report suspicious cases. Such systems suffer from high false-positive rates, undermining their effectiveness and resulting in high operational costs.
We propose a machine learning \emph{triage model}, which complements the rule-based system and learns to predict the risk of an alert accurately. Our model uses both entity-centric engineered features and attributes characterizing inter-entity relations in the form of graph-based features. We leverage time windows to construct the dynamic graph, optimizing for time and space efficiency.
We validate our model on a real-world banking dataset and show how the triage model can reduce the number of false positives by 80\% while detecting over 90\% of true positives. In this way, our model can significantly improve anti-money laundering operations.
\end{abstract}

\section{Introduction}
\label{sec:introduction}

Money laundering concerns the legitimization of criminal proceeds by concealing their origin, resulting in around 2-5\% of global GDP (€1.7-4 trillion) being laundered annually~\cite{lannoo2021anti}. Underlying crimes include drug dealing, human trafficking, fraud, tax evasion, and corruption. Money laundering is, therefore, a severe and global problem affecting people, economies, governments, and the social well-being~\cite{mcdowell2001consequences}.

For financial institutions (FIs), undetected money laundering schemes can result in hefty fines and severe reputational damage. To avoid becoming a vehicle for money laundering, FIs employ compliance experts investigating suspicious behavior. Since it is unfeasible to review all transactions, banks depend upon automated anti-money laundering (AML) solutions to assist investigation teams.

AML solutions typically rely on rule-based systems~\cite{li2017intelligent} to alert suspicious cases based on requirements set by international regulatory agencies, such as the Financial Action Task Force (FATF). Rules are self-explainable, which is often a requirement for auditioners. Investigators subsequently review the alerts and decide whether they are indeed suspicious (i.e., true positives) or false alarms (i.e., false positives). For each true positive, investigators must file a suspicious activity report (SAR). While rule systems have the advantage of interpretability, their simplicity has the drawback of generating many false positives~\cite{weber2018scalable} with reported false positives rates around 95--98\%~\cite{lannoo2021anti}.

\begin{figure*}[t!]
\center
\includegraphics[width=\textwidth]{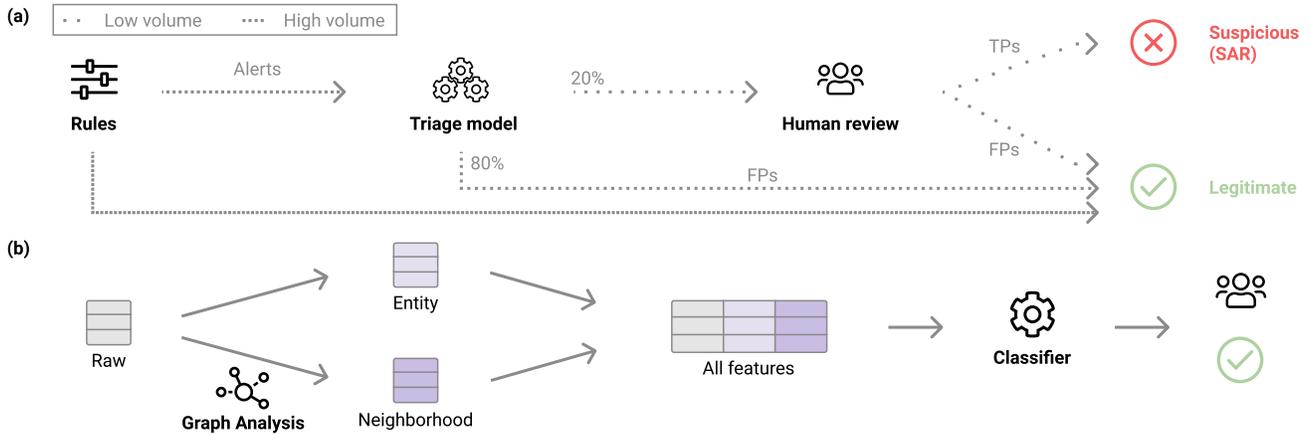}
\caption{Full system overview and triage model details: (a) Our proposed triage component sits downstream of the rules, eliminating the majority of the false positives. (b) Our proposed system leverages various types of features (entity-centric, neighborhood degree and illicit-relations), which are subsequently ingested by our Triage model to classify suspicious activity.}
\label{fig:solution_architectur}
\end{figure*}

In this paper, we aim to reduce false alerts by employing supervised machine learning methods. We call the machine learning component the \emph{triage model}, whose task is to process the alerts generated by the rules (Figure~\ref{fig:solution_architectur}.a). The score of the triage model enables alert suppression (removing low-scoring alerts from the queue to be analyzed) or alert prioritization (ordering the queue of alerts depending on the score). Because all alerts result from the rules, there is no sacrifice in explainability. 

We summarize our main contributions as follows: 
\begin{itemize}
    \item We propose a machine learning component (the \emph{triage model}) that operates on alerts to suppress false positives or prioritize true positives (Figure~\ref{fig:solution_architectur}.a, Section~\ref{subsec:method_classifier}).
    \item We identify promising features to characterize entity behavior (entity-centric) and relationships between entities (graph-based), specifically for the AML domain. These include novel degree-based features as well as a novel version of GuiltWalker~\cite{oliveira2021guiltywalker} features adapted for label delay scenarios. We show how the graphs can be constructed efficiently by keeping a smaller memory for legitimate entities (details in Sections~\ref{subsec:method_profiles},~\ref{subsec:method_graph_features}).
    \item We evaluate our system on a real-world banking dataset in an alert suppression scenario, showing how we can achieve an 80\% decrease in false positives with over 90\% detection of true positives (details in Section~\ref{sec:experiments}). 
\end{itemize}

The remainder of the paper is structured as follows. We first detail our methodology in Section~\ref{sec:method}. Then, we describe the various experiments and discuss results on our real-world dataset in Section~\ref{sec:experiments}. After this, we give an overview of related work in Section~\ref{sec:related_work}. Finally, we present our conclusions in Section~\ref{sec:conclusion}.

\section{Methods}
\label{sec:method}
In  this  section,  we first  provide  an  overview  of our proposed system (section~\ref{subsec:method_classifier}). We then discuss the various features used by our system, namely entity-centric features (section~\ref{subsec:method_profiles}) and neighborhood-centric features (section~\ref{subsec:method_graph_features}).

\subsection{Triage classifier}
\label{subsec:method_classifier}

AML systems typically consist of rule-based systems (illustrated by \textit{Rules} in Figure~\ref{fig:solution_architectur}.a). Such systems are straightforward and interpretable but erroneously alert many legitimate events, overloading human analysts with false positives.
To address this problem, we introduce a novel machine learning component to triage the alerts generated by the rules (\textit{Triage model} in Figure~\ref{fig:solution_architectur}.a). As the model operates on the alerts, it preserves interpretability. The triage model can ingest any features deemed useful for the AML use-case.

\subsection{Profile features}
\label{subsec:method_profiles}

The first set of features engineered for our triage classifier characterize the history of transactions specific to each account. Following \citet{branco2020interleaved}, we call such features \emph{profiles}. Profiles are arithmetic aggregations by a particular field and over a specified time window, such as the total amount spent per account in the past week. These features enable a machine learning model to contrast historical (i.e., long windows) with current behavior (i.e., short windows) and correlate that with suspicious activity. 

In our experiments, we create around 400 profiles features. We consider the account as our grouping entity and aggregate over the amount of money sent and received. We consider five different time windows: one day, one and two weeks, one and two months. The aggregation functions include sum, mean, minimum, maximum, and count. We also compare aggregations over two time windows using ratios and differences. 
Then we apply feature selection using permutation-based feature importance, by training a gradient boosted trees model on a sample of the training dataset, keeping the smallest set of features which cumulatively contribute up to 90\% of the performance for the metric of interest. As a result, we select approximately 100 most important profiles to enrich our data. All profiles are built using our company’s in-house platform, which uses automated machine learning (AutoML) to automatically generate features based on the semantic labels of the data fields (e.g., entity, location, date, or amount)~\cite{marques2020semantic}.

\subsection{Graph neighborhood features}
\label{subsec:method_graph_features}

\begin{figure*}[t!]
\center
\includegraphics[width=\textwidth]{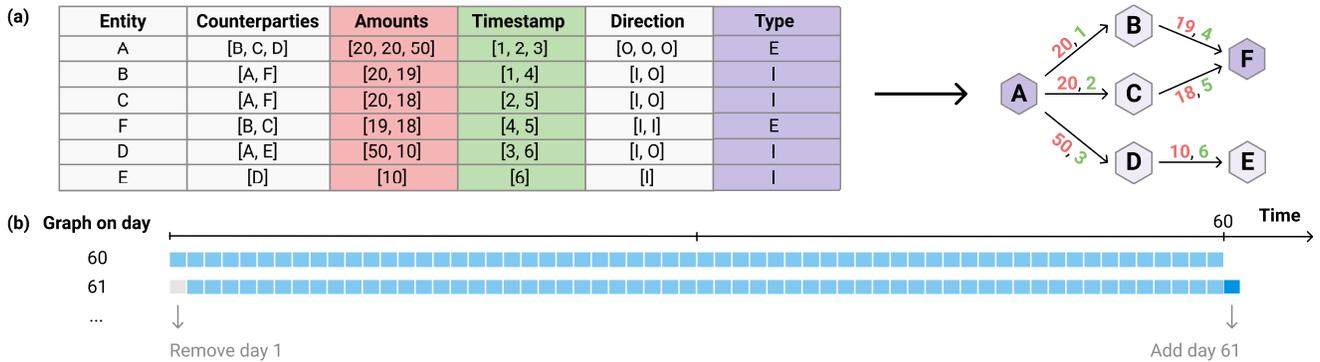}
\caption{Graph construction: (a) Toy example of our tabular data and how we represent it in a graph: each entity (account in this case) is represented by a node, which can have two types (Internal or External). Edges represent transactions between entities (i.e., accounts). Their direction follows the money flow, edges also have the timestamp and amount of a transaction as attributes. (b) shows the sliding window procedure when building the graph, given a time window of 60 days; on the second day, we remove the edges that fall outside the sliding window and add the ones corresponding to the current day.}
\label{fig:graph_construction}
\end{figure*}

\subsubsection{Graph construction}
\label{subsec:method_graph_construction}

Profiles capture entity-centric behavior but ignore inter-entity relations. Hence, we propose to address this shortcoming by enriching the data further with graph features. Because money flows occur between bank accounts, an intuitive choice is to represent accounts as nodes and transactions as edges between accounts. The direction of the edge follows the direction of the money (i.e., from sender to receiver), and edge attributes include the transaction timestamp and amount (Figure~\ref{fig:graph_construction}.a).

Nonetheless, scalability issues arise for financial institutions, often processing millions of events per day and rendering using the entire history unfeasible. Since older events become progressively less relevant, we propose a dynamic graph construction leveraging sliding windows to limit the number of events needed at any moment in time. Similarly to profile features, we compute the graph-based features periodically (every day in our case). We first update the graph snapshot of the previous day by removing the edges that no longer fall within the sliding window and by adding new edges corresponding to the current day's events (Figure~\ref{fig:graph_construction}.b). Secondly, we calculate the graph features for every node with an event in the target day (i.e., accounts requiring evaluation). While the granularity in our experiments is one day, the frequency is adaptable. In the remainder of this section, we discuss various AML specific graph features used by the triage model in our experiments. Nonetheless, the procedure is extensible to include other graph-based features using the same transactional graph in the future.

\subsubsection{Degree features}
\label{subsec:method_degree_features}

We hypothesize that suspicious accounts have more counterparties on average. To convey this information to the model, we calculate the in- and out-degrees of the target node. Additionally, we compute its successors and predecessors' mean, minimum, and maximum in- and out-degrees. In this way, we create eight new features that characterize the number of counterparties of an account and its neighborhood. Analogously, we calculate a \textit{weighted} version of these features by using the transferred amount as the edge weight. 
Because the above features contain information from nodes that are one hop away from the target node, they cannot be captured by the profile features.

\subsubsection{GuiltyWalker features (GW)}
\label{subsec:method_gw_features}

Money laundering patterns typically involve criminal networks. Therefore, suspicious nodes tend to have a higher probability of being connected to other suspicious nodes. This property is captured by GuiltyWalker (GW) features~\cite{oliveira2021guiltywalker} where random walks are generated which stop upon reaching a known illicit node or if there are no available connections. In our implementation, we run 50 random walks for each target node. We then compute the features proposed in the original work, namely features characterizing the length of the random walks (minimum, maximum, median, mean, standard deviation, 25th, and 75th percentile), the fraction of successful random walks (i.e., the "hit rate"), and the number of distinct illicit nodes encountered.

\subsubsection{GuiltyWalker-delay features (GWd)}
\label{subsec:method_gw_with_label_delay}

GuiltyWalker assumes immediate feedback, i.e., that labels are immediately available for all past transactions. In AML, however, investigations are lengthy, resulting in pervasive label delays.
We propose an adaptation of GuiltyWalker by introducing a waiting period. We start by training a machine learning model using profiles and degree features on a first training set. We use the resulting model to score a second training set and define a suitable threshold to obtain \emph{pseudo-labels}. We then compute the GuiltyWalker features using the \emph{pseudo-labels} for the unlabeled transactions in the waiting period and the actual labels otherwise. Finally, we train the triage classifier on the second training set, using profiles, degrees, and GuiltyWalker features.

\section{Results}
\label{sec:experiments}

In this section, we first describe the real-world dataset used in our experiments (Section~\ref{subsec:method_data_set}). We then discuss the experiments with our proposed triage machine learning model, using the raw tabular data enriched with engineered features, namely entity-centric features (Section~\ref{subsec:experiment_profiles}) and neighborhood-centric features (Section~\ref{subsec:exp_graph_features}).

\subsection{Dataset}
\label{subsec:method_data_set}

We use a real-world banking dataset for all our experiments. We can not disclose the bank's identity nor provide exact details for privacy reasons, but we provide approximate metrics to characterize the data where possible.

The raw dataset contains approximately half-million \textit{alerted} transfers between 400,000 accounts and spans over approximately one year. It contains information on whether an account is internal to a bank or external. Transfers occur in both directions between two internal accounts or between an external and an internal account. The dataset is labeled on a transaction level, with a binary label indicating whether a transaction was part of a SAR. However, we devise the proposed triage classifier to generate alerts at the account level, as is typical in AML.
Moreover, in our experiments, we aim to evaluate accounts daily. Hence, we preprocess the raw dataset to contain aggregated daily account features, including total sent and received amounts, the counterparties, the associated timestamps, and the direction. We then extrapolate from the transactional labels to infer the account labels: if there is a suspicious transaction involving an account on a specific day, we mark that account as suspicious on that day. Importantly, this means that suspicious accounts form connected pairs in our preprocessed dataset. The percentage of suspicious accounts in our dataset is less than 3\% of the alerted ones, resulting in over 97\% false positives. The sole categorical feature is account type (external or internal) and encoded with a single binary feature (0 or 1 respectively). Numerical features are kept without further processing. We start from this dataset to compute all subsequent features. 

\subsubsection{Train-val-test split}
We split the dataset temporally into three non-overlapping periods. We use the oldest 60\% for training, then 10\% for validation, and the most recent 30\% for testing our models. For the GuiltyWalker with delay features (see Section~\ref{subsec:method_gw_with_label_delay}), we split the training set further in half.

\subsection{Triage model using entity-centric features}
\label{subsec:experiment_profiles}

All experiments are performed on a real-world banking dataset (Section \ref{subsec:method_data_set}), where the entities to be labeled are the bank accounts. All the models use the same inputs, which are the raw features explained in Section~\ref{subsec:method_data_set}, and a set of around 100 profiles based on each account's sent and received transfer amounts (Section \ref{subsec:method_profiles}). Our first triage model is trained using only the raw features and these entity-centric profile features. 

We aim to maximize the suspicious activity captured by our triage model (true positives) while minimizing incorrect alerts (false positives). We choose our optimization objective to maximize recall at a specific false positive rate (FPR). The FPR can be chosen in accordance with the client. In our experiments, we consider \textit{recall@20\%FPR} as our target metric, which translates to a \textit{reduction} of the false positives by 80\% compared to the rule system itself. Moreover, because most events are legitimate, the chosen FPR (i.e., 20\%) roughly corresponds to the number of alerts to be reviewed to obtain a particular recall. 

We perform an extensive evaluation of various machine learning models, namely Random Forest, Generalized Linear Modeling (GLM), and LightGBM~\cite{ke2017lightgbm} over a wide hyperparameter range. The various algorithms and hyperparameter ranges are reported in Table~\ref{tab:algorithms}.

\begin{table}[!t]
 \begin{center}
\def\arraystretch{1.2}
\begin{tabular}{ |m{2cm}||P{3.5cm}|P{1.8cm}|  }

 \hline
 \textbf{Algorithm}
 &
 \multicolumn{2}{c|}{\textbf{Hyperparameters}} \\
 \hline
 \hline
 \multirow{2}{2cm}{GLM} & Alpha & [0.01 - 0.09]\\
 \cline{2-3}
  & Standardize numericals & [True, False]\\
 \hline
 \multirow{3}{2cm}{Random Forest} & Max depth of trees & [10 - 40]\\
 \cline{2-3}
  & Number of trees & [100 - 200]\\
  \cline{2-3}
  & Min instances for split & [10 - 50]\\
 \hline
 \multirow{3}{2cm}{LightGBM} & Num of leaves & [200 - 500]\\
 \cline{2-3}
  & Min data in leaf & [100 - 200]\\
  \cline{2-3}
  & Learning rate & [0.01 - 0.09]\\
 \hline
\end{tabular}
\end{center} 
\caption{Various algorithms and respective hyperparameter values tried in our hyperparameter search.} 
\label{tab:algorithms}
\end{table}

In total, we trained 50 models on the training set, and Recall\@20\%FPR is compared on the validation set. The top-performing model was a LightGBM, with a test performance close to 80\%recall@20\%FPR. This LightGBM model was considered as our baseline triage model in subsequent experiments.

\subsection{Extending Triage Model with Neighborhood Features}
\label{subsec:exp_graph_features}

We hypothesize that suspicious accounts are differently interconnected than legitimate accounts. To capture such relationships between accounts, we build a directed graph using the accounts as nodes and the transactions between accounts as directed edges (details in Section~\ref{subsec:method_graph_construction}).  Because old events become less relevant for current predictions, we construct the graph dynamically using a sliding window to only include accounts making recent transactions. In the next experiments we create graph snapshots covering time windows of $60$ days as suggested in~\cite{jullum2020detecting}, and subsequently investigate different windows for suspicious and legitimate events.

\subsubsection{Degree Features.} 
\label{subsec:experiment_degrees}
We first hypothesize that the number of neighbors and the money flow may differ based on the account class. Therefore, we calculate the in- and out-degree features of a node and its one-hop neighbors. The neighbor degrees are aggregated using mean, minimum, and maximum operations (details in Section~\ref{subsec:method_degree_features}). Adding these neighborhood degree features to the baseline model improves the performance by 11.6 percentage points (p.p.) \textit{Recall@20\%FPR}, corroborating our initial hypothesis. (Figure~\ref{fig:results_delta_recall_graph_features}, \emph{+Degrees}).
We also calculate \textit{weighted} versions of these features, where the weight is the amount of money being transferred, but did not surpass the performance of standard degree features (Figure~\ref{fig:results_delta_recall_graph_features}, \emph{+Weighted Degrees}).

\subsubsection{GuiltyWalker features.}
\label{subsec:experiment_gw}
\begin{figure}[t!]
\center
\includegraphics[width=0.49\textwidth]{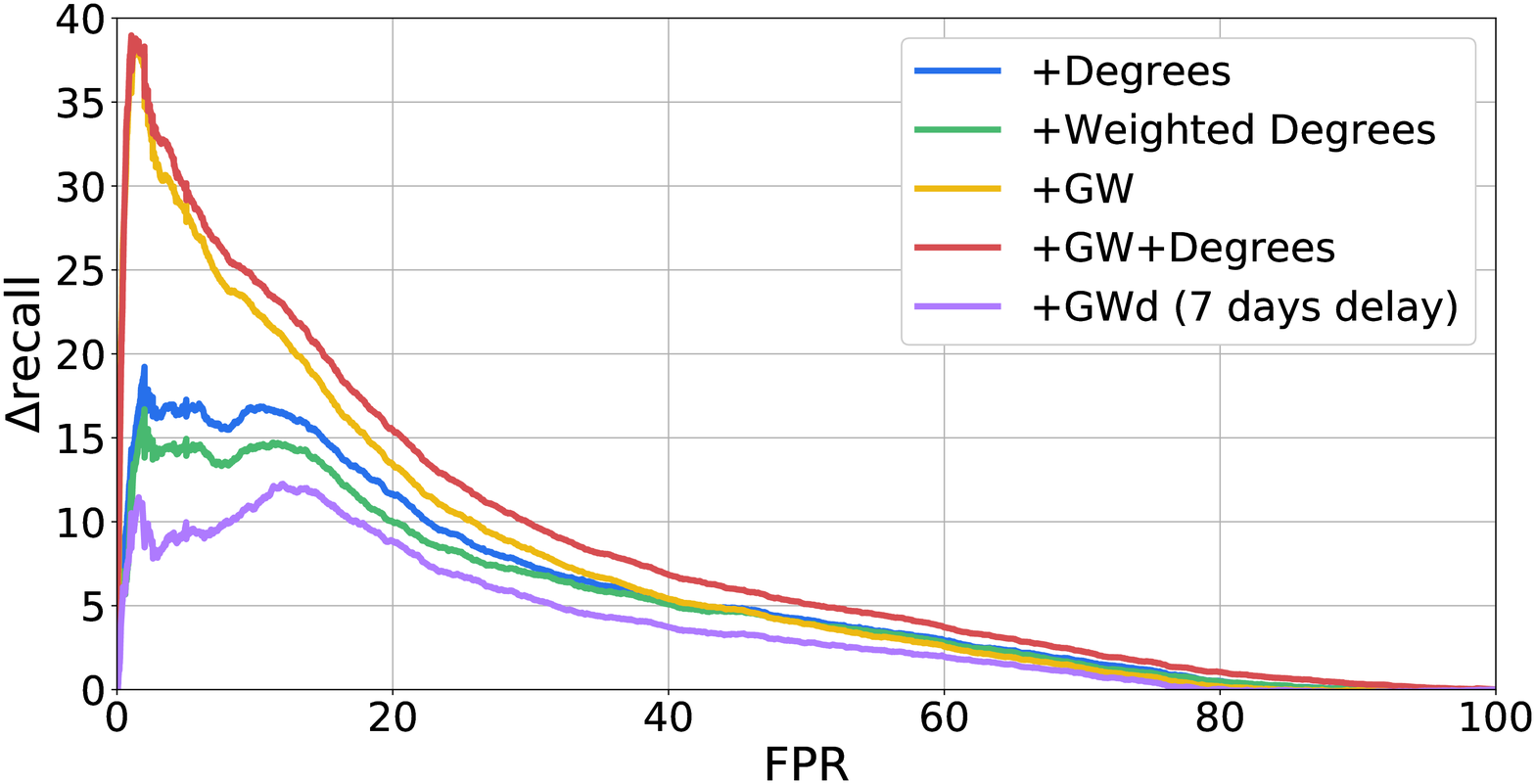}
\caption{Triage models with graph features: Difference in recall compared to our baseline model without graph features. GuiltyWalker (GW) features achieve large gains at low FPR. Adding degrees on top of GW features further improves performance at intermediate FPRs (GW+Degrees).}
\label{fig:results_delta_recall_graph_features}
\end{figure}

Because money laundering often manifests itself in networks of criminal accounts, we hypothesize an increased probability of finding connected suspicious nodes. We, therefore, compute GuiltyWalker (GW) features~\cite{oliveira2021guiltywalker}, which capture distance to suspicious nodes using random walks (see Section~\ref{subsec:method_gw_features}). Adding the GW features to the baseline model improves the performance by 13.4 p.p. in \textit{Recall@20\%FPR} (Figure~\ref{fig:results_delta_recall_graph_features}, \emph{+GW}). Interestingly, GW improves performance up to 38\% in the lower FPR region.

We subsequently add the degree features and the GuiltyWalker feature in conjunction with the baseline model and investigate whether they capture different information. Our results show that the two sets contain overlapping information, but the combined model is still better than models using any of the features separately: the gain achieved is 15.5 p.p. \textit{Recall@20\%FPR} (Figure~\ref{fig:results_delta_recall_graph_features}, \emph{+GW+Degrees}).

\begin{figure}[t!]
\center
\includegraphics[width=0.49\textwidth]{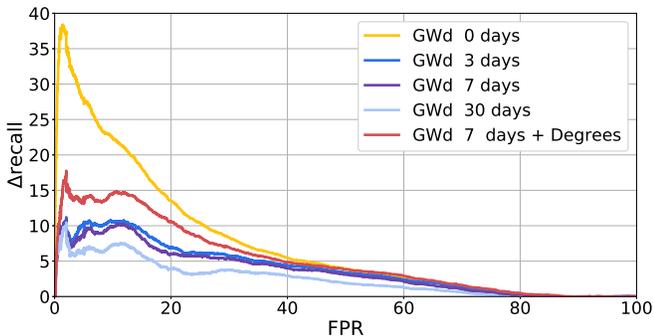}
\caption{Label delay experiments: Increasing delay decreases recall. Adding degree features boosts the performance of the GWd models (shown for seven days delay: GWd 7 days + Degrees).}
\label{fig:results_delta_recall_label_delay}
\end{figure}

\subsubsection{GuiltyWalker-delay features (GWd).}
Our experiments show that GW features capture useful information for our task of reducing false positives. However, the original GW algorithm~\cite{oliveira2021guiltywalker} assumes that every day we have all the accurate labels from the previous days. This assumption does not apply in realistic banking AML scenarios because it takes time for analysts to escalate alerts, review cases, and file SARs. We, therefore, adapt the GW algorithm to leverage a model score and a threshold to create pseudo-labels instead of the actual label for the most recent events (details in Section~\ref{subsec:method_gw_with_label_delay}). To determine the threshold hyperparameter, we perform a grid search using the following values: 0.1, 0.15, 0.25, 0.5, 0.89 (a threshold that maintains the same percentage of positives in the validation set as in the train set), and 0.43 (a threshold corresponding to 20\%FPR when evaluating the validation set using our baseline model). Assuming a label delay of 7 days, we found that the gain in recall@20\%FPR achieved a peak of 6.4 p.p. for a threshold of 0.25 while performing worse for smaller or larger thresholds.

We experiment with label delay scenarios ranging from 1 day to 30 days. As expected, the shorter the labeling delay, the better the performance (Figure~\ref{fig:results_delta_recall_label_delay}). Despite a drop in performance compared to the original GW features at low FPRs, an exciting outcome of this experiment is that GWd features are beneficial when added to the baseline model, even with a label delay of one month.

We then test the combination of degree features and GWd features with our baseline model. While adding the degree features improves the detection of true positives, we notice that it does not beat results when using the degree features without GWd (compare blue line in Figure~\ref{fig:results_delta_recall_graph_features} with the red line in Figure~\ref{fig:results_delta_recall_label_delay}). We verify that this is also true when degrees are calculated on the smaller training set used for the GWd models (data not shown). Because the degrees features are computationally lighter to compute, a good choice in scenarios like ours is discarding the GWd features.

\subsubsection{Sliding window and memory of suspicious nodes}
\label{subsec:methods_window}
\begin{figure}[t!]
\center
\includegraphics[width=0.49\textwidth]{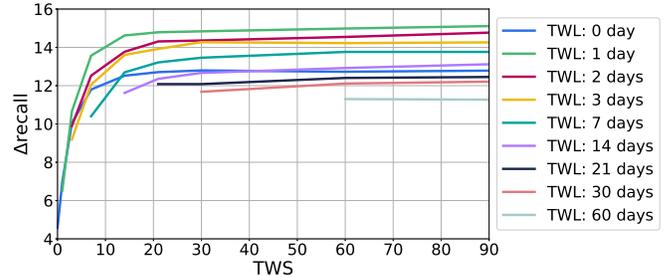}
\caption{Time window experiments: Time windows for suspicious events (TWS) reach a plateau in performance for values of 20-30 days. Time windows for legitimate events (TWL) reach maximum performance for a value of 1 day.}
\label{fig:results_window_sizes}
\end{figure}

In the previous experiments, we built a dynamic graph using a sliding time window of 60 days (Section~\ref{subsec:method_graph_construction}). We now wondered how changing this window affects the triage model performance. Moreover, since our experiments showed that connections to known suspicious accounts are important features, we investigate whether keeping a more extended memory for such suspicious accounts compared to legitimate ones is helpful. To this end, we use a different window for each event type (legitimate vs. suspicious) and retrain our best model for a realistic case of label delays, which uses only degrees features as described in the previous section. Similar results were obtained when retraining the best model without label delay (using degrees+GW features, data not shown). We perform a grid search for values of these time windows between 0 days (i.e., events are not used in the graph at all) up to 90 days. For brevity, we refer to the time window for legitimate events as TWL and to the time window for suspicious events as TWS. Firstly, we find that, for any value of TWS, the best performance is achieved for a TWL equal to one day. Secondly, the performance increases only marginally when increasing the TWS beyond 30 days (Figure~\ref{fig:results_window_sizes}). Therefore, we can construct a good model efficiently by keeping only one day of legitimate events and 30 days of suspicious events in our graph. Importantly, having separate time windows for legitimate and suspicious events implies knowing the label at least after the duration of the smallest time window. Thus, for a label delay of 7 days, the best model we can construct efficiently would be using a TWL of 7 days and a TWS of 30 days. Nonetheless, it is interesting that we can significantly reduce the data needed to construct the graph without sacrificing performance.

\section{Related Work}
\label{sec:related_work}

The most common systems to detect money laundering based on transaction data employed by banks are rule based~\cite{weber2018scalable}. In this work, we tackle the issue of reducing false positives of such systems, therefore keeping the rules for compliance and explainability reasons. In contrast, most of the existing works discussed below replace the rule-based system entirely and try to tackle two problems at once: reducing false positives on the one hand and reducing false negatives on the other hand. For thorough reviews of machine learning approaches for AML, we refer the reader to~\cite{chen2018machine,tiwari2020review}.

We can divide the machine learning AML systems into unsupervised and supervised methods. The majority use unsupervised techniques due to the lack of real-world labeled datasets available in the money laundering domain. The typical approach is to firstly cluster events, followed by anomaly detection. To address the lack of data, various strategies have been proposed. Either a fully synthetic dataset is generated~\cite{luna2018finding,drezewski2012system}, or only unusual accounts are simulated within a real-world dataset~\cite{tang2005developing,liu2008sequence,gao2009application,wang2009research}, or one assumes that rare events within a peer group are suspicious~\cite{larik2011clustering}. One drawback of anomaly detection approaches is the assumption that suspicious activities are outliers, which may not always be the case since money launderers try to simulate legitimate behavior~\cite{lorenz2020machine}. Arguably, better validations of the systems were reported in~\cite{yang2014dbscan,camino2017finding,shokry2020counter} using analyst feedback, or in~\cite{liu2010scan} using real labeled data and where authors report a 52\% recall@5\%FPR.

Supervised methods leverage labeled training data. Luo et al.~\cite{luo2014suspicious} generate synthetic data and propose a classification algorithm based on association rules to detect suspicious events.
Other researchers use real-world datasets and aim to detect suspicious behavior by training classification algorithms like SVM~\cite{keyan2011improved} where authors report 64\% recall@6\%FPR, XGBoost~\cite{jullum2020detecting} obtaining an AUC of 82\%, or after comparing various algorithms~\cite{zhang2019machine} in which the best model was a neural network and obtained 74\% AUC.
The performances of various models are hard to compare across the studies due to their different metrics and datasets.

Recent work has tried to incorporate graph information in the AML system in order to capture network patterns. Weber et al. 
\citet{weber2019anti} benchmarked graph convolutional networks against various supervised methods and concluded that random forest algorithms provide a better performance, despite the lack of graph-based information. Oliveira et al.~\cite{oliveira2021guiltywalker} propose GuiltyWalker, leveraging random walks on a cryptocurrency graph to characterize distances to previous suspicious activity. The authors reported a 5 p.p. improvement in F1 score when including these novel features. Random walks were also used in~\cite{hu2019characterizing} on top of a transaction graph representing the bitcoin network. \citet{savage2016detection} propose a community detection approach, from which neighborhood-centric features are extracted and ingested by a supervised machine learning model. On a real-world dataset, the best model was a random forest classifier achieving over 80\%recall@20\%FPR.

Finally, other works propose graph-based suspiciousness scores based on money flows~\cite{li2020flowscope,sun2021cubeflow}. These algorithms do not use a learning algorithm and instead build a detection system incorporating business knowledge about money flows. The scope is to detect novel types of money laundering activity (i.e., reducing false negatives), while our goal is to reduce incorrectly alerted events (i.e., reducing false positives).

\section{Conclusion}
\label{sec:conclusion}

In this work, we proposed a machine learning triage model to reduce false positives of an AML rule-based system. Our triage model is deployed \emph{after} the rule-based system, only processing alerted events and therefore not replacing the rule-based system. In this way, maintain the explainability and compliance of the rules.

The triage model is trained on a real-world banking dataset enriched with two sets of engineered features, entity-centric features capturing the usual entity behavior and graph-based neighborhood features capturing the characteristics of entity interactions. We show how both sets of features lead to substantial improvements, and combining them allows us to alert only 20\% of the events while capturing over 90\% of the suspicious entities. Moreover, for time and memory efficiency, our graph is computed using sliding windows, and we show how it is beneficial to keep a longer memory of the suspicious accounts than the legitimate ones. Because the majority of accounts are legitimate, this significantly reduces the computational load on the system.

In this work, we experimented with two types of graph-based features (degree and GuiltyWalker), but our framework can be easily extended in the future to include new graph-based features. Moreover, if relevant in the respective dataset, the proposed neighborhood degree features could be expanded to include more distant node information. Our novel version of the GuiltyWalker features (GWd) works in a realistic AML setting with label delay. We showed how this delayed version remains beneficial for our classifier, despite a drop in performance compared to the non-delayed version. However, this benefit is abolished if degree features are included, which therefore contain overlapping information in our dataset. Moreover, the GWd performance depends on the power of our baseline model, and improving this model in the future may close the gap between the delayed version and the original algorithm.

We have evaluated all experiments using recall@20\%FPR. This metric was chosen to have an alert suppression strategy in mind, i.e., discarding alerts that receive a small triage model score. The same setup can also be used in an alert prioritization strategy, using the triage model score directly to order the alerts for review, or a hybrid strategy combining suppression and prioritization. Ultimately, using our triage model in any such situation will improve anti-money laundering systems.

\bibliography{references}

\end{document}